\documentclass[letterpaper]{article} 
\usepackage{aaai25}  
\pdfoutput=1
\usepackage{times}  
\usepackage{helvet}  
\usepackage{courier}  
\usepackage[hyphens]{url}  
\usepackage{graphicx} 
\usepackage{natbib}  
\usepackage{caption} 
\usepackage{algorithm}
\usepackage{algorithmic}
\usepackage{url}
\usepackage{comment}
\usepackage{array}
\usepackage{subfigure}
\usepackage{amssymb}
\usepackage{amsmath}
\usepackage{multirow}
\usepackage{multicol}
\usepackage{color}
\usepackage{tabularx}
\usepackage{booktabs}
\usepackage{colortbl}
\usepackage{makecell}
\usepackage{pifont}
\newcommand{\cmark}{\checkmark}%
\newcommand{\xmark}{$\times$}%
\usepackage{newfloat}
\usepackage{listings}
\DeclareCaptionStyle{ruled}{labelfont=normalfont,labelsep=colon,strut=off} 
\floatstyle{ruled}
\newfloat{listing}{tb}{lst}{}
\floatname{listing}{Listing}
\title{PoseTalk: Text-and-Audio-based Pose Control and Motion Refinement for One-Shot Talking Head Generation}

\author{
    Jun Ling\textsuperscript{\rm 1}\equalcontrib\thanks{Project Leader}, Yiwen Wang\textsuperscript{\rm 1}\equalcontrib, Han Xue\textsuperscript{\rm 1}, Rong Xie\textsuperscript{\rm 1}, Li Song\textsuperscript{\rm 1,2}\thanks{Corresponding Author}\\
}
\affiliations{
    \textsuperscript{\rm 1}Institute of Image Communication and Network Engineering, Shanghai Jiao Tong University, Shanghai, China\\
    \textsuperscript{\rm 2}Cooperative Medianet Innovation Center, Shanghai Jiao Tong University, Shanghai, China\\
    \{lingjun, song\_li\}@sjtu.edu.cn
}

\usepackage{bibentry}

\begin{document}

\maketitle

\begin{abstract}

While previous audio-driven talking head generation (THG) methods generate head poses from driving audio, the generated poses or lips cannot match the audio well or are not editable. 
In this study, we propose \textbf{PoseTalk}, a THG system that can freely generate lip-synchronized talking head videos with free head poses conditioned on text prompts and audio. The core insight of our method is using head pose to connect visual, linguistic, and audio signals. 
First, we propose to generate poses from both audio and text prompts, where the audio offers short-term variations and rhythm correspondence of the head movements and the text prompts describe the long-term semantics of head motions. To achieve this goal, we devise a Pose Latent Diffusion (PLD) model to generate motion latent from text prompts and audio cues in a pose latent space. 
Second, we observe a loss-imbalance problem: the loss for the lip region contributes less than 4\% of the total reconstruction loss caused by both pose and lip, making optimization lean towards head movements rather than lip shapes. 
To address this issue, we propose a refinement-based learning strategy to synthesize natural talking videos using two cascaded networks, \textit{i.e.}, CoarseNet, and RefineNet. The CoarseNet estimates coarse motions to produce animated images in novel poses and the RefineNet focuses on learning finer lip motions by progressively estimating lip motions from low-to-high resolutions, yielding improved lip-synchronization performance. 
Experiments demonstrate our pose prediction strategy achieves better pose diversity and realness compared to text-only or audio-only, and our video generator model outperforms state-of-the-art methods in synthesizing talking videos with natural head motions. \underline{Project: \url{https://junleen.github.io/projects/posetalk}}.

\end{abstract}

\section{Introduction}
\label{sec:intro}

One-shot audio-driven talking head synthesis aims to generate a talking video based on a single-face image and input speech. In recent years, it has garnered significant attention from the community due to its massive potential in the industry, including virtual assistants, digital humans, virtual conferencing, and the like. Benefiting from large-scale talking face datasets~\cite{zhang2021flow,kaisiyuan2020mead}, numerous approaches have been proposed, demonstrating rapid progress in this field~\cite{ling2020toward,tian2024emo}. 

\begin{figure}[t]
  \centering
  \includegraphics[width=0.99\linewidth]{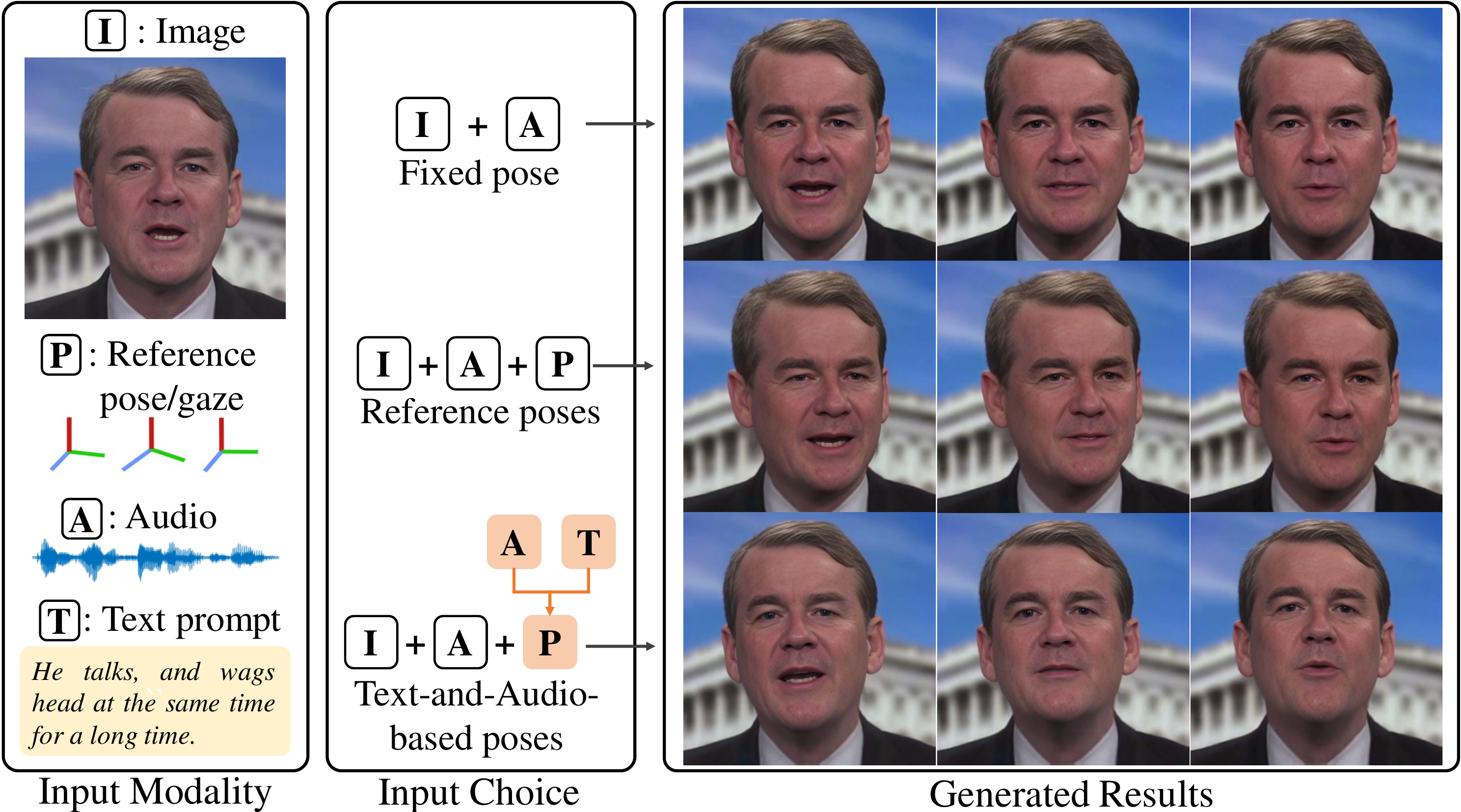}
  \caption{Key features of our PoseTalk. Our method can synthesize talking face videos from an image, the driving audio, and driving poses (In the following sections, we use ``\textit{pose}" as the abbreviation of ``\textit{pose/gaze}" for readability). The driving poses might be fixed, reference poses (from other talking videos or predicted from audio), or generated poses based on text prompts and audio. Our approach additionally supports the generation of diverse poses using different text prompts. Due to the page constraints, we present these results in our supplementary materials and demonstration videos. }
  \label{fig:teaser}
\end{figure}

Existing one-shot talking head generation methods can be categorized into template-based and audio-driven. 
Template-based methods reconstruct face videos by inpainting mouth features based on the mouth-masked face images~\cite{prajwal2020lip,cheng2022videoretalking,zhong2023iplap,shen2023difftalk,stypulkowski2024diffused} or generate facial images driving by the pose references from off-the-shelf videos and audio features~\cite{zhou2021pose,liang2022expressive,wang2022pdfgc}. 
Although these methods produce lip-sync results, the borrowed poses are not easy to modify and do not always match the latent dynamics and rhythm of the input audio. 
On the contrary, audio-driven methods generally predict head poses conditioned on the input audio~\cite{zhang2023sadtalker,wang2021audio2head,xu2024hallo}, and then synthesize the talking face video conditioned on the input audio and predicted poses. Audio-driven methods have been receiving increasing attention due to their possibility of generating talking videos from a single image. 

Despite their utility, audio-driven methods might be limited by two \textbf{D}rawbacks: \textbf{D1}: Audio-driven methods often learn pose prediction based on a short-term audio feature segment without context head motion hints. However, short-term audio feature segments lack long-term semantics of the head motions and are insufficient for faithful head pose generation. For example, a boy might either wag his head or remain still with only his mouth moving while reading. This challenge might make the predicted poses rigid and converge to mean pose sequences (\textit{e.g.}, waging around the mean pose)~\cite{ye2023geneface}.
\textbf{D2}: While all approaches take image reconstruction loss or lip-sync loss to penalize the errors between the generated frames and ground-truth frames or speech content, the losses are not equally considered. For example, users are generally more sensitive to errors in small-scale lip motions (\textit{e.g.}, lip motions are not synchronized with audio segments) and the naturalness of large-scale head motions. However, it has been reported that the loss for lip region contributes less than 4\% of the total reconstruction loss~\cite{prajwal2020lip}, making the optimization lean towards head poses rather than lip shapes. As a result, the model usually focuses on optimizing head poses before generating fine-grained lip motions.

We address the above issues from two perspectives. 
To tackle \textbf{D1}, we introduce a Pose Latent Diffusion (PLD) model to achieve pose generation conditioned on both audio and text prompts describing actions. 
Text prompts are a user-friendly interface to express human intention and they can offer long-term cues for head movements, while the audio provides short-time information (such as rhythm or sentiment, slight nodding when someone says ``yes''). 
Based on our analysis, we devise a latent diffusion model that leverages both conditions to predict pose sequences. Specifically, we first train a VAE and compress an original pose sequence into a latent embedding. Then we train a conditional diffusion model to learn correlations between text-audio and poses in the motion latent space, followed by a VAE decoder to decode the pose sequence from the predicted motion latent. 
Through this approach, text and audio can collaborate effectively in facilitating the learning of natural head movements. 

To address the loss-imbalance problem \textbf{D2}, we find that lip-oriented motion refinement is the key to achieving accurate lip motion prediction and pose transfer. 
To this end, we propose a two-stage motion prediction framework: the pose-oriented coarse motion prediction stage and the lip-oriented motion refinement stage. In the first stage, we train a pose-oriented coarse motion network to predict coarse motions and achieve pose transfer. In the second stage, we devise a lip-oriented motion refinement network to compensate for the inadequacy of coarse motion. Specifically, we borrow the strategy from~\cite{wang2022latent} and build multi-resolution motion flows, progressively predicting finer lip motions from low-resolution to high-resolution and finally formulating fine-grained lip motions (as illustrated in Fig.~\ref{fig:liprefine_net}). Compared to the one-stage prediction, the proposed two-stage approach exhibits more advantages in learning accurate motions. In addition, we find that the generated images in the second stage might lose some details such as the backgrounds. To alleviate this issue, we adopt the same structure as the editing network of PIRender~\cite{ren2021pirenderer} conditioning on refined motion results and the source frame (Please refer to supplementary material for details). 

To summarize, our major contributions are in four folds: 
\begin{itemize}
    \item We introduce text prompts and audio conditions to talking face generation and propose a novel and efficient system to achieve fine-grained pose control and audio-driven talking face video generation (shown in Fig.~\ref{fig:teaser}). 
    \item We propose a pose latent diffusion model, conditioned on text prompts and audio, to achieve user-friendly head pose generation (shown in supplementary materials). 
    \item We show that employing refinement-based motion prediction for talking video generation has the potential to strike a balance between the accurate animation of head movements and synchronized lip motion generation. 
    \item We conduct extensive experiments to show the capability of our approach in generating high-fidelity and lip-synchronized audio-driven talking videos, proving the effectiveness of our contributions. 
\end{itemize}

\section{Related Work}
\label{sec:related_works}

\subsection{Audio-Driven Talking Head Synthesis}

Recent works on one-shot talking head generation can be roughly categorized into template-based methods and audio-driven methods. Template-based methods~\cite{prajwal2020lip,stypulkowski2024diffused,shen2023difftalk,cheng2022videoretalking,zhong2023iplap} bypass the pose editing and focus exclusively on synthesizing the lip movements from speech and background images. However, the downside of these methods is the inability to control poses, coupled with the limitation to mouth movements only. Some methods~\cite{zhou2021pose,zhou2019talking,liang2022expressive,wang2022pdfgc} adopt the pose details from reference videos. Although effective, the references pose might not fit the input audio thus undermining video realness. Audio-driven methods predict head poses from audio only~\cite{zhang2023sadtalker,ma2023dreamtalk,zhang2021flow,ma2023styletalk}. A common practice often leverages predefined 3DMM expression and pose coefficients to describe lip motions and head motions, and then learns those coefficients from audio features~\cite{ren2021pirenderer,xue2022high}. 
Likewise, few works~\cite{zhou2020makelttalk,chen2019hierarchical,ji2022eamm,Gan2023efficientEAT} predict intermediate mouth representations, \textit{e.g.}, facial landmarks, keypoints, or latent motion weights. However, the 3DMM coefficients and the facial landmarks are sometimes ineffective in describing subtle lip movements and might compromise lip motion accuracy in real-world applications. 

Recently, the diffusion model has shown promising achievements in image/video generation~\cite{rombach2022high,ho2022video,guo2023animatediff} and talking face generation~\cite{he2024gaia,xu2024hallo,tian2024emo}. 
GAIA~\cite{he2024gaia} introduces a latent diffusion model to learn the correspondence between speech and motion latent in the latent space. Hallo~\cite{xu2024hallo} and EMO~\cite{tian2024emo} introduce temporal modeling to the Stable Diffusion~\cite{rombach2022high} model and optimize the denoising UNet to generate talking videos and improve temporal smoothness. Although promising in high-fidelity image synthesis, these methods compromise the training and inference speed and require huge amounts of high-quality videos (over 200 hours). 
Meanwhile, researchers also develop methods on person-specific tasks~\cite{li2023ernerf,guo2021adnerf,yang2023context,ling2023stableface,tang2022memories}. However, these methods cannot deal with arbitrary facial images. In this work, we develop a method that can effectively generate talking head videos to fulfill the requirement of motion descriptions and enhance the synthesis of full facial expressions more naturally and seamlessly.

\subsection{Motion Generation}
We treat the poses as a part of human motions and present related advances in recent years. 
Motion generation is a fundamental task in computer vision. It allows various inputs of multi-modal data such as text, action, and images. 
Earlier explorations such as~\cite{wang2021audio2head,chen2020talking} predict head motions from audio using LSTM-based temporal modules. SadTalker~\cite{zhang2023sadtalker} employs a VAE-based~\cite{kingma2013autovae} model to model pose predictions conditioned on audios. Recently, text-to-motion has stood out among conditional motion generation tasks thanks to human language's user-friendly and understandable nature. The essence of the text-to-motion generation is learning a shared latent space for language and motion. Typical methods~\cite{guo2022generating,petrovich2022temos} focus on learning the correlations between the input texts and the motion latent through attention modules and transformers. 
MotionCLIP~\cite{tevet2022motionclip} aligns the motion latent space with the text and image spaces of the pre-trained visual language model CLIP~\cite{radford2021learning}, expanding text-to-motion beyond data limitations. 
UDE~\cite{zhou2023ude} discretizes motion sequences into latent codes, predicts quantized codes using a transformer similar to GPT, and decodes motions via a diffusion model decoder. MoFusion~\cite{dabral2023mofusion} employs a diffusion model with a 1D U-Net-like transformer to reconstruct motion sequences from natural language or audio inputs. MDM~\cite{tevet2023humanmdm} employs a classifier-free guided diffusion model to predict samples instead of noise at each diffusion step, capable of adapting to different conditional modes. MLD~\cite{chen2023executing} maintains a diffusion process in the latent motion space instead of using a diffusion model to establish connections between raw motion sequences and conditional inputs. However, compared to the human motion generation task, synthesizing head pose involves considering not only the user's text-driven input but also the corresponding audio, which is another key factor influencing the effectiveness. Previous methods have focused more on either text-guided or audio-guided approaches, neglecting the inherent connection between these two forms of guidance.

\section{Method}
\label{sec:method}

Our generation framework consists of two key modules: pose generation and video generation. The pose generation module utilizes a latent diffusion model to learn motion representations from text descriptions and audio within the VAE latent space. The video generation module employs a refinement-based generator to synthesize realistic talking head videos based on audio inputs and the generated poses. In the following sections, we elaborate on our data construction pipeline, as well as the pose generation model and refinement-based video generator.

\begin{figure}[!t]
  \centering
  \includegraphics[width=0.95\linewidth]{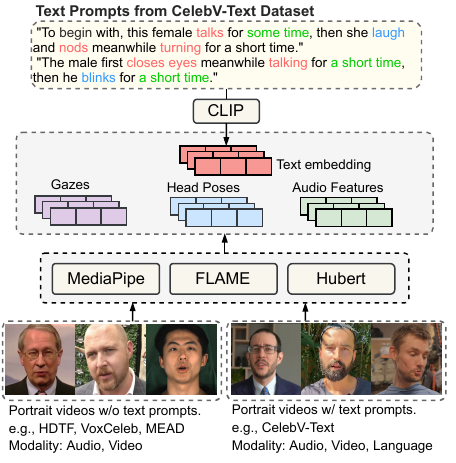}
  \vspace{-0.2cm}
  \caption{Dataset construction pipeline. We adopt off-the-shelf vision models and audio-pertaining models to extract motion-related representations and audio features, respectively. Then, for the text prompts that describe head movements from the CelebV-Text dataset, we adopt the text encoder from CLIP~\cite{radford2021learning} to obtain semantic-aligned text embeddings. }
  \label{fig:dataset_preparation}
\end{figure}

\subsection{Dataset Construction}
To learn models about both pose prediction and video generation, large-scale paired samples, especially the paired sample of motion description, audio, and head poses, are required. However, most of the existing talking video datasets, such as VoxCeleb~\cite{nagrani2017voxceleb} and HDTF~\cite{zhang2021flow}, have no paired action descriptions. On the other hand, the facial text-video dataset such as CelebV-Text~\cite{yu2023celebv-text} is not appropriate for audio-driven face synthesis because of large pose angle and the audio might be background music, noise, or from the invisible speakers. To alleviate this issue and fully use the videos of different datasets, we employ 6DoF head poses and audio as motion-related representations to build connections across different datasets. 

We formulate two categories of datasets: 1) the dataset contains (\texttt{audio}, \texttt{pose}, \texttt{text}) for training pose prediction model; and 2) the dataset of lip-synced talking videos (\texttt{audio}, \texttt{pose}, \texttt{video}) for training the talking video generation. For the first category, we collect samples from the CelebV-Text dataset which contains 700000 in-the-wild face videos along with text descriptions of head actions. To clean the data, we removed the videos that have extreme head poses or noisy audio. Finally, we obtained 60000/4000/800 samples for training/testing/validation. For the latter category, we collect videos from three talking video databases, \textit{i.e.}, MEAD~\cite{kaisiyuan2020mead}, HDTF, and VoxCeleb. We sampled videos from both frontal view and left30$^\circ$/right30$^\circ$ views in the MEAD dataset and finally obtained 12060 talking videos. We sampled 377 videos from HDTF for training and the rest 30 videos for testing. In addition, we followed~\cite {siarohin2019first} and collected 18000+ talking videos from VoxCeleb for training. 

The dataset construction pipeline is shown in Fig.~\ref{fig:dataset_preparation}, where we estimate the head poses using a 3DMM model~\cite{feng2021deca}, predict the gazes through MediaPipe~\cite{lugaresi2019mediapipe}, and extract the sequential audio features by using Wav2Vec2.0~\cite{baevski2020wav2vec} and toolkit~\cite{ott2019fairseq}. 
We utilize CLIP to extract text embedding for the CelebV-Text dataset.

\subsection{Pose Generation}
\label{subsec:text2pose_generation}
\begin{figure}[t]
  \begin{minipage}[a]{1.0\linewidth}
    \centering
    \includegraphics[width=1.01\linewidth]{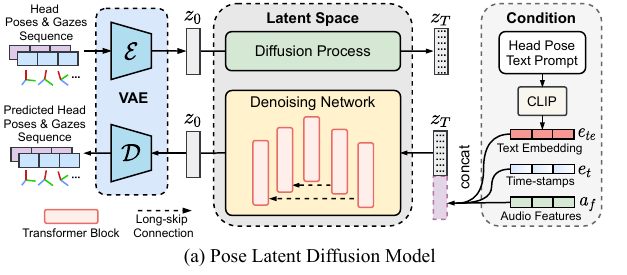}
  \end{minipage}
  \vspace{0.2cm}
  \begin{minipage}[b]{1\linewidth}
    \centering
    \includegraphics[width=0.99\linewidth]{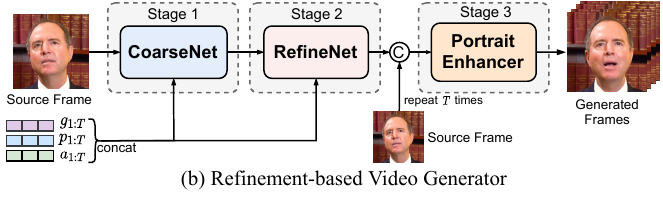}
  \end{minipage}
  \vspace{-1.cm}
  \caption{The overview of our pose diffusion and talking face video generation. (a) During training, the pose latent diffusion model is conditioned on the pose embedding learned by VAE. The denoising process is conditioned on the text embedding, time stamps, and audio features. (b) Given a source image, the audio features, and the extracted or predicted pose/gaze features, the video generator gradually estimates finer motions and lip-synced talking videos. }
  \label{fig:text-and-audio2pose}
\end{figure}

The diffusion model has been applied to data generation in many areas, especially motion generation, speech synthesis, and video synthesis. However, diffusing directly on raw pose sequences is inefficient and requires significant computational resources. To reduce this demand, we first utilize a Variational AutoEncoder (VAE), $\mathcal{V}=\{\mathcal{E}, \mathcal{D}\}$, to learn a low-dimensional latent space for diverse sequences of head poses and eye gazes. 
Our VAE is illustrated in Fig.~\ref{fig:text-and-audio2pose}-a(left). The encoder $\mathcal{E}$ is employed to extract representative features $z \in \mathbb{R}^{n \times d}$ from the original sequences $x^{1:T}$ and the decoder $\mathcal{D}$ could reconstruct the latent into motion sequences $\hat{x}^{1:T}$. We use HuberLoss to train our VAE model:
\begin{equation}
    L_{\mathrm{VAE}}=\mathrm{HuberLoss}(x^{1:T},\hat{x}^{1:T})
\end{equation}

Following MLD~\cite{chen2023executing}, we build a Transformer-based denoising model with long skip connections which is more suitable for sequential data. The diffusion process is performed on a representative and low-dimensional motion latent space of our trained VAE model. Given the generated latent space features, we can reconstruct the pose sequence through the decoder.

In the Pose Latent Diffusion module (PLD), both text and audio guide the generation process. The text condition provides a general guide to actions, while the audio condition offers more detailed information, such as rhythm, which helps to specialize the results. This combination combines global guidance with detailed adjustments, offering the sequence more semantic guidance. For text conditions, the CLIP \cite{radford2021learning} text encoder is employed to map text prompts for its powerful semantic understanding capabilities. For audio conditions, we extract features from the audio using~\cite{baevski2020wav2vec}. By combining both text embedding and audio features, our approach generates more natural yet action-controllable pose sequences. The conditional objective can be written as follows:
\begin{equation}
L_{\mathrm{PLD}}:=\mathbb{E}_{\epsilon, t, c}\left[\left\|\epsilon-\epsilon_\theta\left(z_t, t, c)\right)\right\|_2^2\right]
\end{equation}
where $t$ denotes the timestep, $z_t$ represents the noisy feature, and $c$ represents the condition comprimising text and audio.

During inference, we employ the classifier-free guidance \cite{ho2022classifier} as follows:
\begin{equation}
\epsilon_\theta^s\left(z_t, t, c\right)=s \epsilon_\theta\left(z_t, t, c\right)+(1-s) \epsilon_\theta\left(z_t, t, \varnothing\right),
\end{equation}
where $s$ is the guidance scale and set to 7.5 to enforce the impacts of guidance.

\begin{figure*}[ht]
  \centering
  \includegraphics[width=1\linewidth]{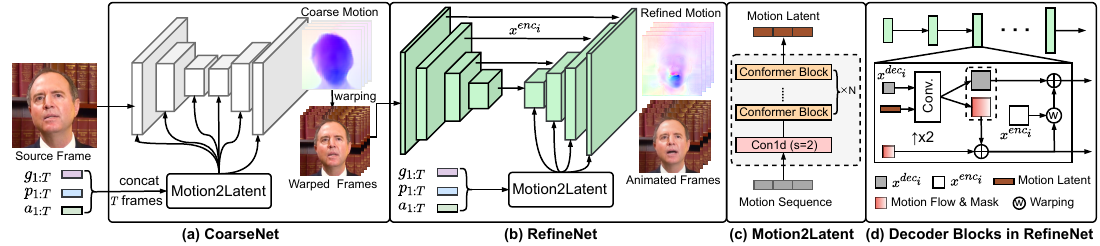}
  \vspace{-0.3cm}
  \caption{Refinement-based Video Generator. (a) We employ a CoarseNet to estimate coarse motions based on the source frame and input conditions. Motion2Latent module (see c) is used to map the motion $m_{1:T}$ to latent space. In (b), we utilize an image encoder to extract features from the warped frames and develop an effective motion refinement decoder to progressively output refiner lip motions from low-to-high resolutions. (d) The structures of blocks in Motion Decoder of RefineNet. }
  \label{fig:liprefine_net}
\end{figure*}

\subsection{Refinement-based Video Generator}
\label{subsec:cascaded_motion_refinement}
Due to the challenge of striking a balance between large-scale head motions (\textit{e.g.}, head rotations and translation) and small-scale motions (\textit{e.g.}, lip movements) in a single network, we divide the motion prediction process into two stages to alleviate this issue. 
As shown in Fig.~\ref{fig:text-and-audio2pose}-b, our video generator consists of three submodules, a CoarseNet that focuses on pose transfer, a RefineNet that gradually refines lip motions from low to high resolution, and a Portrait Enhancer which improves the final generation quality. The three models are sequentially trained in three stages. 

\noindent\textbf{CoarseNet.}
\label{subsec:pose_coarse_motion}
In the first stage, we train a coarse motion prediction model (CoarseNet) to focus on pose transfer and optimize using global reconstruction loss. To learn the motions from the input source frame and the motion condition, we design two networks: 1) The Motion2Latent network is employed to produce a latent vector of $z$ from the concatenated motion conditions; 2) An image-to-motion prediction network that predicts coarse motions from source frames and the motion latent. 
Motion smoothness is crucial for generating face videos, and the dynamic correlations among context motions are also influenced by past features. To this end, we treat the input as a sequence and model context information and dependencies using a Motion2Latent network. The Motion2Latent network is a Conformer-based~\cite{Gulati2020conformer} architecture, which involves a 1d-convolution layer with stride 2 and 5 conformer blocks. The details of Motion2Latent are shown in Fig.~\ref{fig:liprefine_net}-c. Using the projected motion latent, we employ an Hourglass-like network and insert the motion latent to the features through the AdaIN~\cite{Huang2017arbitrary} layer after each convolution layer. Given the output coarse motion, we employ differentiable backward warping operation to warp the source image and obtain the warped frames.

\noindent\textbf{RefineNet.}
\label{subsec:iterative_motion_refinement}
The RefineNet progressively refines the lip shapes and motions based on the outputs of frozen CoarseNet, compensating for the limitation of coarse motion prediction models. As we care more about the motion cues provided by audio features than the pose details in the refinement stage, we employ another Motion2Latent network to fuse the input motion parameters. 

Inspired by the success of multi-resolution motion estimation~\cite{wang2022latent}, we build our motion refinement on a multi-scale and low-to-high resolution progressive prediction scheme. First, we utilize an encoder to extract the multi-scale features (denoted by $x^{enc}$) from the warped frames. For a better understanding, in the $i$-th layer of the Motion Decoder, we denote the layer block by $Conv._{i}$ and the input features by $x^{dec_{i-1}}$, and the output flow and mask by $d\mathcal{F}_i$ and $m_i$. We upsample the motions from lower resolution twice and denote it as $\mathcal{F}^{\uparrow}_{i-1}$ and add it to the flow at $i$-th layer. Then, we warp the encoder feature using the updated flows. This process can be written as:
\begin{equation}
\begin{split}
    \mathcal{F}_{i} := d\mathcal{F}_{i}+\mathcal{F}^{\uparrow}_{i-1}& ,\ m_i:= m_i + m^{\uparrow}_{i-1} \\
    x^{out_i}=warp(x^{enc_i}, \ \mathcal{F}_{i})&\cdot m_i + x^{dec_i}(1-m_i),
\end{split}
\end{equation}
where $x^{enc_i}/x^{dec_i}$ is the encoder$/$decoder feature at the $i$-th layer, and $\mathcal{F}^{\uparrow}_{0}$ is the output of first convolution in Fig.~\ref{fig:liprefine_net}\-d. Through the motion updating process within the motion decoder, we obtain refined motions from the resolution of $8\times8$ to $256\times256$. The final motion flow from the input image to the target image can be formulated as:
\begin{equation}
    \mathcal{F}_i = \mathcal{F}_{coarse} + d\mathcal{F}_{i} + d\mathcal{F}_{i-1}^{\uparrow} + d\mathcal{F}_{i-2}^{\uparrow\uparrow} + \cdots, 
\end{equation}
where $\mathcal{F}_{coarse}$ is the coarse motion we obtained through the coarse motion network. To obtain the final output image, we use a convolution layer and a Tanh activation layer to output three-channel images from deformed features.

\noindent\textbf{Portrait Enhancer.}
As previously described, our two-stage motion refinement process has successfully achieved pose editing on a large scale and synthesis of fine-grained mouth movements. However, the quality of the generated images still has some imperfections which are mainly caused by the warping operations and the synthesis process in the second stage. To better recover facial details and maintain image consistency, we trained an image enhancer using a UNet structure from PIRender~\cite{ren2021pirenderer}. This enhancer takes as input the source frame and the animated frames from the second stage to produce the final result.

\noindent\textbf{Training Losses.}
\label{subsubsec:training_losses}
We adopt the multi-scale perceptual loss~\cite{siarohin2019first} $\mathcal{L}_{perc}$ and lip-sync loss~\cite{prajwal2020lip} $\mathcal{L}_{sync}$ to train our video generation model, both in stage 1 and stage 2. To implement lip-sync loss, we use differentiable backward warping to align the faces in the center of the image and calculate synchronization loss on the generated images and audio segments. In stage 3, we optimize the Portrait Enhancer using only the perceptual loss. We elaborate on the details in the supplementary material.

\section{Experiments}
\label{sec:experiments}
\subsection{Implementation Details}
\label{subsec:Training}

\begin{table*}[!t]
    \centering
    \small
    \setlength{\tabcolsep}{1.25mm}
    \begin{tabular}{l|ccccc|ccccc|ccc}
    \toprule
     \multirow{2}{*}{Method}   &   \multicolumn{5}{c|}{HDTF~\cite{zhang2021flow}}    &   \multicolumn{5}{c|}{MEAD~\cite{kaisiyuan2020mead}} &    \multicolumn{3}{c}{Control Pose via} \\
      & SSIM$\uparrow$ & LPIPS$\downarrow$  &   IP$\uparrow$ & AED$\downarrow$ &   Sync$_{\rm conf}$$\uparrow$   & SSIM$\uparrow$ &  LPIPS$\downarrow$ &  IP$\uparrow$  & AED$\downarrow$  &   Sync$_{\rm conf}$$\uparrow$ &   Audio  &   Text  & Video \\
     \hline
     Wav2Lip~\shortcite{prajwal2020lip}  &  0.801   &  0.156  &  \underline{0.855}   & \underline{0.1792} &   \underline{7.78}  &  0.768  & 0.145 & 0.832 & \underline{0.1738}  &   \textbf{8.17} & \xmark   &   \xmark   &   \cmark \\
     PC-AVS~\shortcite{zhou2021pose}   &  0.546  &   0.392  &  0.520  & 0.1942  &  6.33  &  0.485 & 0.440  & 0.541 &  0.1834  & 5.97  & \xmark  & \xmark   &   \xmark \\
     SadTalker~\shortcite{zhang2023sadtalker}  &  0.646 &  0.207 &  0.838  &  0.2048  & 5.53   &  0.662  & 0.144  &  0.875 & 0.1766  &  5.36   & \cmark  &  \xmark &   \cmark \\
     IP\_ LAP~\shortcite{zhong2023iplap}    &  \textbf{0.827}   &   \underline{0.112}     &  0.819  &  0.1807 &   5.18   & 0.813  &  0.131  &  0.865  &   0.1812    & 4.79  & \xmark   &   \xmark   &   \cmark \\
     PD-FGC~\shortcite{wang2022pdfgc}  &  0.580   &    0.249   &    0.491   &  0.2020 &   7.10    & 0.683     & 0.235   &  0.411 &    0.1988  &  \underline{7.12} & \xmark   &   \xmark   &   \cmark  \\
     Hallo~\shortcite{xu2024hallo} &    -   &   -   & \textbf{0.861}  & 0.1799  &  7.75 &  -    &   -   & 0.865  &  0.1770 & {7.03} &   \cmark  &  \xmark   &   \xmark  \\
     PoseTalk (Ours)    & \underline{0.815}  &  \textbf{0.110} & 0.850 & \textbf{0.1763}  &  \textbf{7.89}   &  \textbf{0.832} &  \textbf{0.097}  & \textbf{0.897}  &  \textbf{0.1640}   & {7.11}  & \cmark  &    \cmark  &   \cmark \\
    \bottomrule
    \end{tabular}
    \vspace{-0.2cm}
    \caption{Quantitative comparisons with state-of-the-art methods on MEAD~\cite{kaisiyuan2020mead} and HDTF~\cite{zhang2021facial}. Our method consistently achieves better performance than previous approaches. Meanwhile, our approach better aligns the synchronization score to ground truth videos. Our approach also enables more flexible input conditions, including audio, pose reference or text-based motion descriptions. The best and second best results are marked in \textbf{bold} and \underline{underline}, respectively. }
    \label{table:compare_with_sota}
\end{table*}

The training process of PLD consists of two stages: training the VAE and training the diffusion model. Our VAE is trained for 100 epochs with a batch size of 16 and our diffusion model is trained for 100 epochs with a batch size of 128.
We train the video generation model in three stages. In the first stage, we train the CoarseNet on audio-visual datasets for 200k iterations with a batch size of 20. Then, we deliberately train the RefineNet using audio features and pose guidance for 400k steps while freezing the CoarseNet. Finally, we train the Portrait Enhancer to recover the missing details in generated images for 200k steps. To improve the model's robustness to mouth positions, we randomly crop the face images in the dataset. 

Following general evaluation protocols~\cite{zhong2023iplap,chen2023executing}, we conduct evaluations on video generation and pose generation, respectively. 
In addition, we also conducted a user study to assess the generated talking videos which are generated using unseen text prompts and driving audio. The results and essential implementation details are presented in supplementary materials.

\subsection{Evaluation of Video Generation}
\label{subsec:exp_results}

\begin{figure}[t]
    \includegraphics[width=1\linewidth]{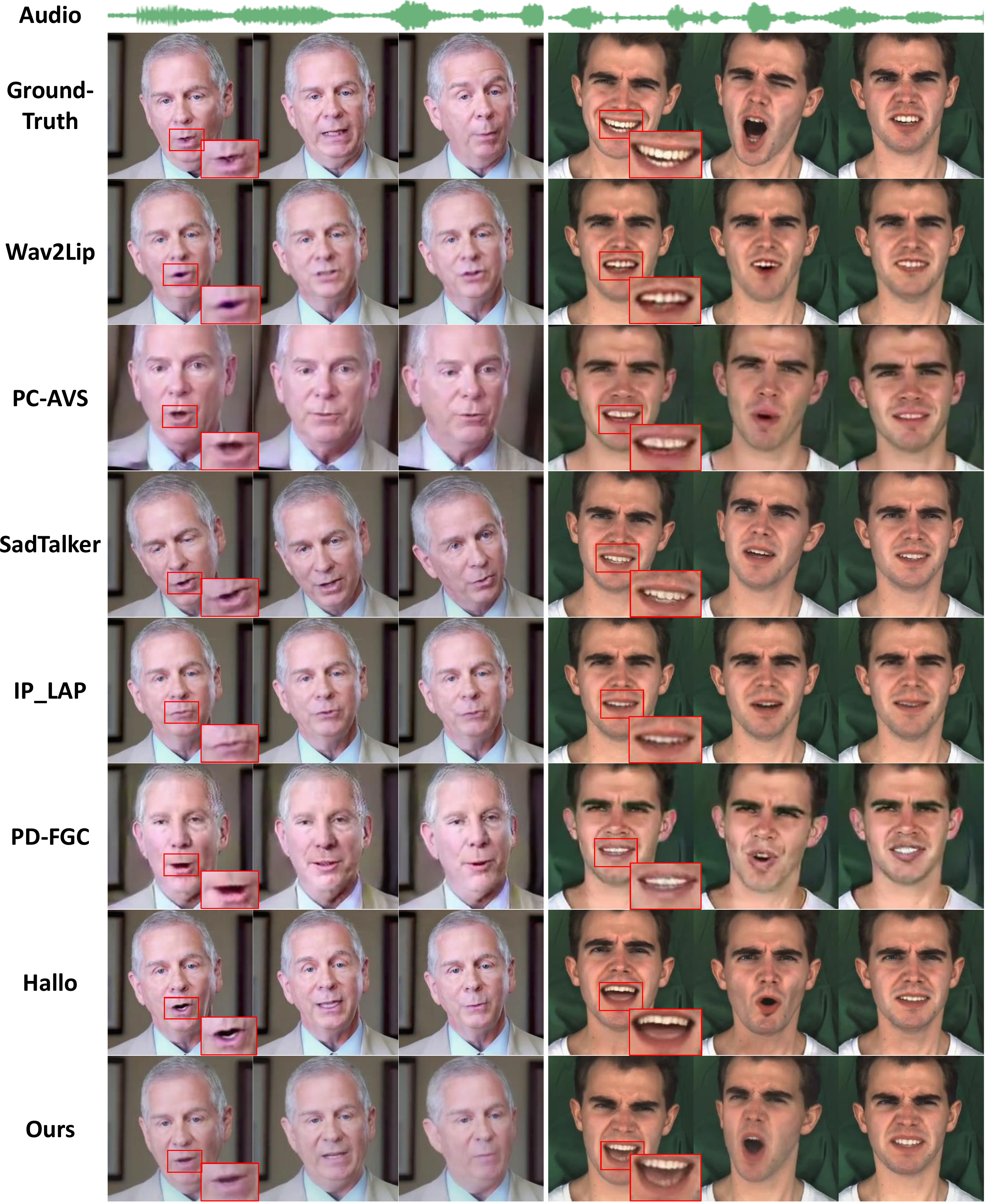}
    \vspace{-0.4cm}
    \caption{Qualitative comparisons with state-of-the-art methods on HDTF and MEAD. We obtain the results under the one-shot audio-driven generation settings. }
    \label{fig:compare_with_sota}
\end{figure}

\noindent\textbf{Metrics.}
We adopt LPIPS~\cite{zhang2018unreasonable} and SSIM~\cite{wang2004image} to assess the image reconstruction quality. 
We also adopt the Identity Preserving (IP) score to assess cosine identity similarity between the source and the generated images using the backbone of Arcface~\cite{deng2019arcface}. 
It is worth noting that the resolutions and mean positions of the generated images and ground truth images differ from method to method, making it challenging to evaluate mouth landmarks distance for different methods. To combat this, we adopt the Average Expression Distance (AED)~\cite{ren2021pirenderer} to measure the average 3DMM expression distance between the generated and real images. In addition, we adopt Sync$_{\rm conf}$~\cite{chung2016out} score to evaluate the audio-visual synchronization quality. 

\noindent\textbf{Face Video Generation Baselines.} We compare our method with several representative works, including the template-based methods (Wav2Lip~\cite{prajwal2020lip}, IP\_LAP~\cite{zhong2023iplap}), reference-based methods (PC-AVS~\cite{zhou2021pose}, PD-FGC~\cite{wang2022pdfgc}), and audio-driven method (SadTalker~\cite{zhang2023sadtalker}, Hallo~\cite{xu2024hallo}). Template-based methods have some advantages on reconstruction quality as they directly `copy' the backgrounds and only synthesize the mouth region. As reference-based methods require a reference video to provide pose details, we then use the original video as the pose reference for these methods. To compare with SadTalker, we extract head poses from original frames or directly predict pose parameters from audio depending on whether the audios are paired with the source frame.

\begin{figure*}[t]
    \centering
    \includegraphics[width=0.95\linewidth]{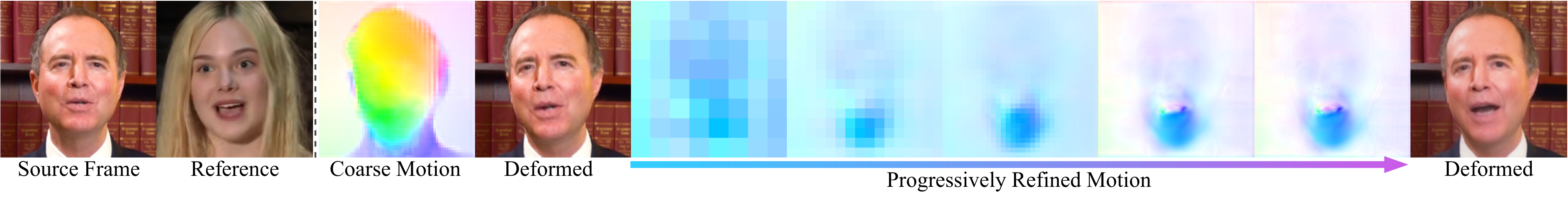}
    \vspace{-0.2cm}
    \caption{Visualization of the coarse motion and the refined motion, we show refined motion flows from resolution 16$\times$16 to 256$\times$256. Compared to the coarse motion, the refined motions can capture finer details of the lip motions in higher resolutions. }
    \label{fig:refined_motion_visualization}
\end{figure*}

\subsubsection{Comparison with State-of-the-art methods.}
In Table~\ref{table:compare_with_sota}, we show the testing results on HDTF and MEAD. 
Wav2Lip attains the best score in Sync$_{conf}$ as it was optimized on SyncNet. Compared to existing methods, our approach shows better overall generation quality and motion accuracy. 
Next, we conduct the qualitative comparison and show the results in Fig.~\ref{fig:compare_with_sota}. As can be observed, Wav2Lip tends to produce blurry results and distorted inner-mouth textures, which further sacrifice the overall video realness. PC-AVS heavily relies on face alignment operation to ensure that the faces are strictly centered, which introduces severe head jitters and frame inconsistency. SadTalker animates the source image using the predicted expression coefficients, which are insufficient to capture the subtle deviations of lip motions. IP\_LAP and PD-FGC tend to produce face images with either blurry results or texture distortions. Among these methods, our approach achieves comparable or better performance in expression similarity and lip-sync accuracy. Hallo generates high-resolution face images utilizing the pertaining models of Stable Diffusion Models~\cite{rombach2022high}, which in turn, requires much more computational resources and time to generate the same video samples as our method.

\begin{table}[!t]
    \centering
    \setlength{\tabcolsep}{1mm}
    \small
    \begin{tabular}{l|ccccc}
    \toprule
     Method     &  SSIM$\uparrow$  &  LPIPS$\downarrow$     &   AED$\downarrow$   & Sync$\rm _{conf}$$\uparrow$   \\
     \hline
     CNet. w/o $\mathcal{L}_{sync}$  & 0.785  &   0.121  &  0.2003  &   4.15   \\ 
     CNet. w/ $\mathcal{L}_{sync}$   & \underline{0.804}  &  0.119 &  0.1925  &  4.68   \\
     CNet. + RNet. w/o $\mathcal{L}_{sync}$   &  0.801   &  \underline{0.112}  & \underline{0.1792}  &   \underline{6.73}    \\
     Full Model & \textbf{0.815}  &  \textbf{0.110}    &  \textbf{0.1763}  &  \textbf{7.89}   \\
    \bottomrule
    \end{tabular}
    \vspace{-0.1cm}
    \caption{Results of ablation studies on the motion refinement strategy. CNet.: CoarseNet, RNet.: RefineNet.}
    \label{tab:ablation_two_stage_motion}
\end{table}

\noindent\textbf{Ablation of Motion Refinement Strategy.} 
We investigate the motion refinement strategy in our designs and show results in Table~\ref{tab:ablation_two_stage_motion}. Specifically, we test three variants concerning the motion refinement module and lip-sync loss. For example, we remove the motion refinement module and train the CoarseNet without (or with) the synchronization loss $\mathcal{L}_{sync}$ (the first row and the second row). Then, we remove the lip-sync loss in the refinement stage to investigate its influence (third row). As can be inferred, the generation model benefits from $\mathcal{L}_{sync}$ and this loss improves the lip-sync quality while incorporating both RefineNet and $\mathcal{L}_{sync}$ attains much better results compared to other variants. 

\begin{table}[!htbp]
    \centering
    \small
    \setlength{\tabcolsep}{1mm}
    \begin{tabular}{l|c|cc|cc|ccc}
    \toprule
    Input   & GT    &   \multicolumn{2}{c|}{LSTM} & \multicolumn{2}{c|}{Transformer} & \multicolumn{3}{c}{PLD}  \\
    \hline
    Text    & -   & $\times$&    \checkmark  &$\times$&    \checkmark  &$\times$ &    \checkmark &   \checkmark   \\
    Audio   & - &   \checkmark &   \checkmark  &   \checkmark &   \checkmark  & \checkmark &   $\times$ & \checkmark   \\
    \hline
     FID $\downarrow$   &  0    & 6.54 & 24.15 & 10.55 & 28.64  & 7.43  &   \textbf{5.32}    & \underline{5.88}  \\
     Div. $\uparrow$    & 15.87 & 5.48 & 4.47 & 6.62 & 5.87     & \underline{13.16} &   12.16   & \textbf{15.77} \\
     \bottomrule
    \end{tabular}
    \vspace{-0.1cm}
    \caption{Quantitative results on pose generation guided by different input conditions and generation backbones. We adopt the LSTM-based pose prediction module from Audio2Head~\cite{wang2021audio2head} and the transformer-based network~\cite{chen2022transformers2a}. }
    \label{table:pose-related}
\end{table}

\begin{figure}[!ht]
  \centering
    \includegraphics[width=0.99\linewidth]{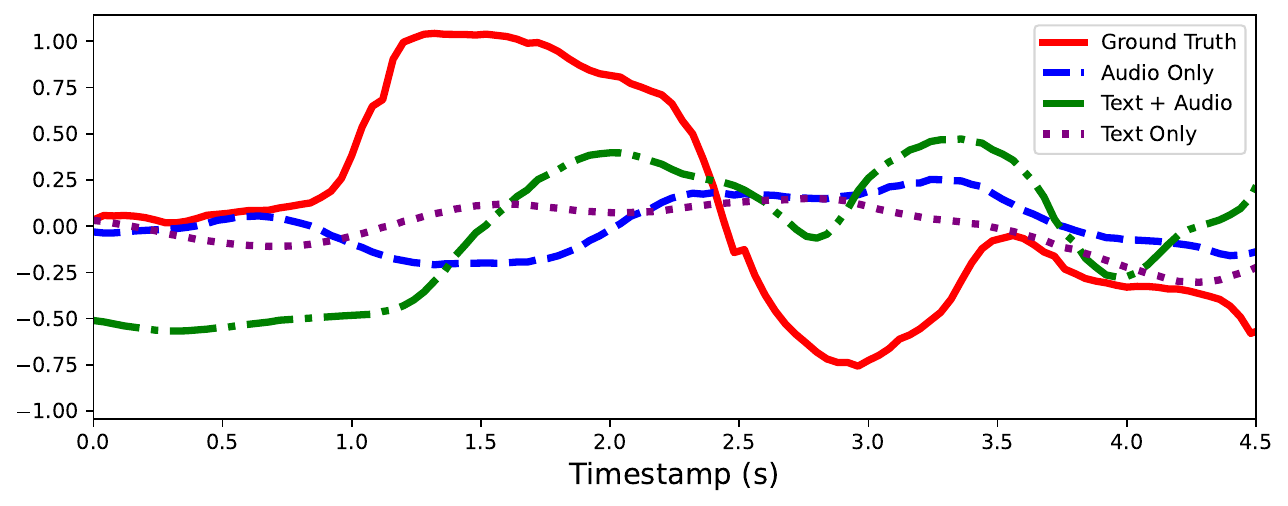}
    \vspace{-0.1cm}
    \caption{Motion curve of the predicted head poses.}
    \label{fig:motion_curves_beat}
\end{figure}

\begin{figure}[!ht]
    \includegraphics[width=0.99\linewidth]{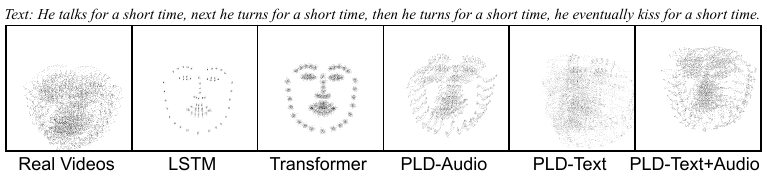}
    \vspace{-0.1cm}
    \caption{Landmark tracemaps of the generated videos. }
    \label{fig:landmarks_tracemap}
\end{figure}

\subsection{Ablation of Pose Latent Diffusion}
Here we conduct experiments to evaluate our PLD model. Specifically, we explore model performance w.r.t different input types (text-only, audio-only, and text-and-audio) and generation backbones (LSTM-based, transformer-based, and diffusion-based). 
We adopt Frechet Inception Distance (FID)~\cite{heusel2017gans} to evaluate generation quality by measuring the feature distribution distance between the generated and real motions. Meanwhile, we assess the diversity (Div.) of the generated motion sequence by calculating the variance of features~\cite{guo2022generating}. 
The comparison results are shown in Table~\ref{table:pose-related}. As can be found, audio or text is not enough to produce highly diverse motion sequences. Although the FID scores slightly increased when involving both audio and text guidance, the generated sequences became more diverse and rhythmic, leading to clear improvements in the diversity score. For better visualization, we reduce the head motions into 1 dimension using PCA, and present the sequential results in Fig.~\ref{fig:motion_curves_beat}. The pose curve of full guidance has the closest variation to the ground truth as it captures more details in both the overall long-term semantics and the finer local motions. 
Using the same text prompt and audio, we predict head poses and animate a face image to obtain talking face videos. Then, we estimate the per-frame facial landmarks and present the tracemaps as Fig.~\ref{fig:landmarks_tracemap}. As can be found, Audio2Head and Transformer show subtle head movements. Using only PLD-Text might result in large-scale head movements, which are improper for talking heads. Among these choices, our PLD-Text+Audio produces the most similar results as real videos.

\section{Conclusion}
This paper presents a systematical solution for text- and audio-based talking video generation. Our approach integrates two effective components: a pose latent diffusion model and a refinement-based video generator, addressing pose prediction and video generation, respectively. 
By learning the latent diffusion model on disentangled poses, our framework not only predicts natural head movements from action prompts and audio but also allows user-engaged and flexible head pose control. 
Additionally, we propose a refinement strategy to relieve the burden of the loss-imbalance issue and improve lip-synchronization quality for talking video generation. 
We hope our perspective will broaden avenues for future research in this area.

\begin{links}
    \link{Project}{https://junleen.github.io/projects/posetalk}
\end{links}

\bibliography{aaai25}

\setcounter{secnumdepth}{2} 
\appendix
\vspace{1cm}
\noindent{\LARGE \textbf{Supplementary Material}}
\vspace{0.2cm}

\noindent This supplementary file presents detailed information on the architecture, training losses, metrics, and more experimental results. 

\section{Implementation Details}

\subsection{Inference}
\label{subsec:inference_pipeline}
\begin{figure}[ht]
    \centering
    \includegraphics[width=0.69\linewidth]{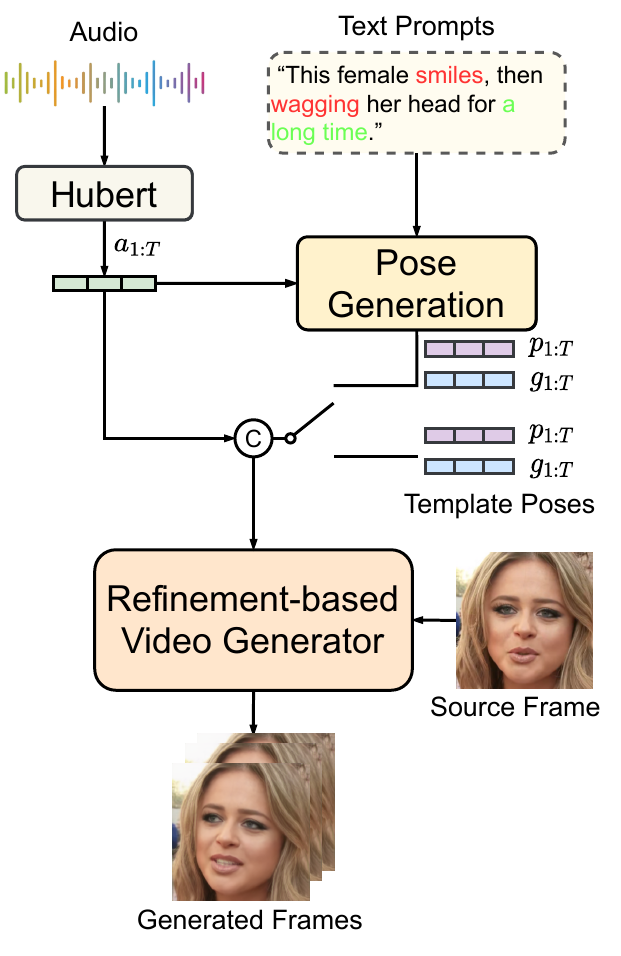}
    \vspace{-0.3cm}
    \caption{Our inference pipeline consists of two modules: 1) The Pose Generation module generates diverse poses guided by the audio features and text prompts. 2) The Refinement-based Video Generator synthesizes lip-synchronized talking videos given the input audio features, poses, and source frame. It is worth noting that our poses can also be sampled from template poses. }
    \label{suppfig:inference_pipeline}
\end{figure}

Our method is capable of generating natural talking face videos from a single source image, input audio, and text-based pose descriptions. 
During training, we train a pose latent diffusion model (PLD) and a refinement-based video generator for the pose and video generation separately, using poses as intermediate representations to connect these two tasks. 
During inference, the PLD predicts natural pose sequences from text descriptions and input audio. Then, we employ the video generation model to synthesize realistic talking videos from the audio features and the generated pose sequence. The inference pipeline is shown in Fig.~\ref{suppfig:inference_pipeline}. 
As can be seen, our refinement-based video generator can be driven by the poses which are either deferred from input audio and text prompts or sampled from templates. Such property ensures model usage is more flexible and user-friendly than prior methods.

\begin{figure*}[t]
    \centering
    \includegraphics[width=0.95\linewidth]{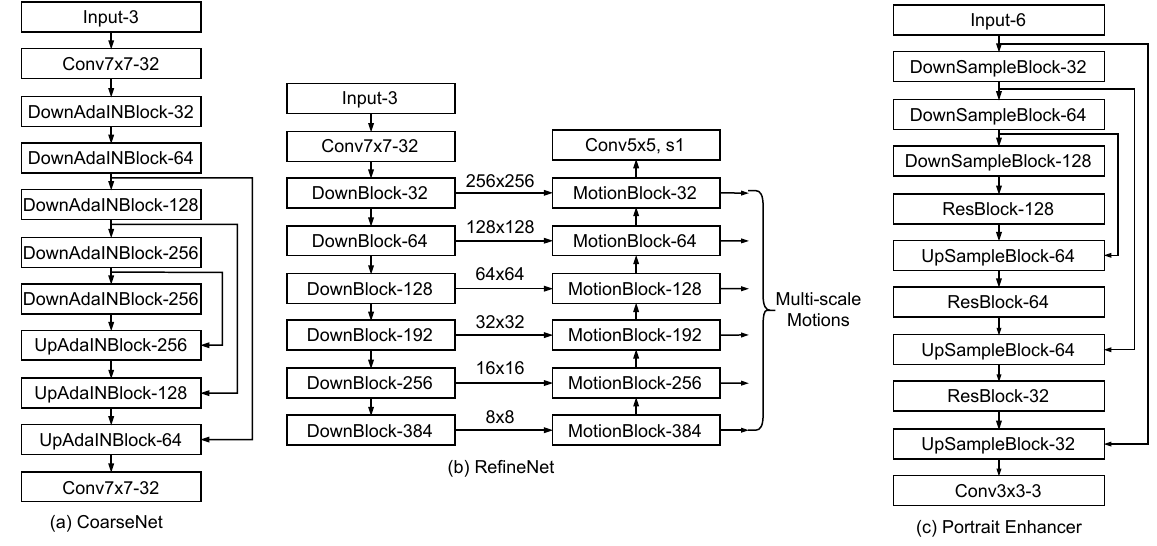}
    \vspace{-0.3cm}
    \caption{The network architectures of the CoarseNet, RefineNet, and Portrait Renderer. }
    \label{suppfig:network_architecture}
\end{figure*}

\begin{figure*}[t]
    \centering
    \includegraphics[width=1\linewidth]{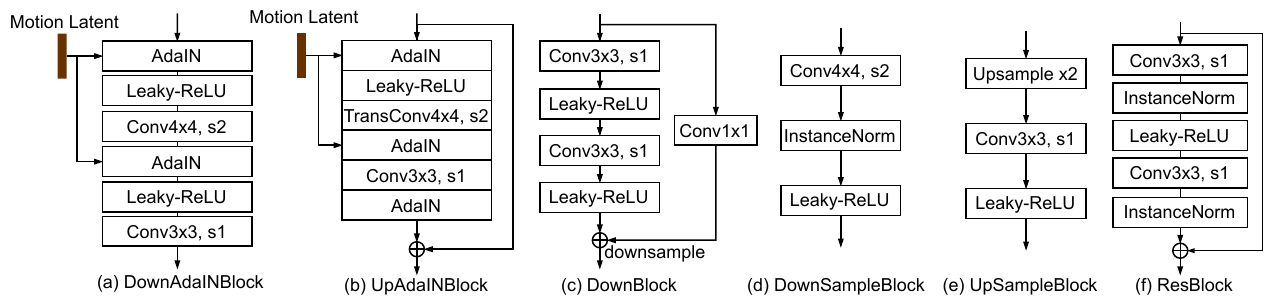}
    \vspace{-0.5cm}
    \caption{The network components employed in our networks. }
    \label{suppfig:network_blocks}
\end{figure*}

\subsection{Architecture Details}
Our system mainly consists of a CoarseNet for pose-oriented motion prediction, a RefineNet for lip motion refinement, and a Portrait Enhancer for details enhancement. As mentioned in our main manuscript, the CoarseNet is implemented using U-Net-based architecture, with AdaIN~\cite{Huang2017arbitrary} layer to fuse the motion latent with image features. The network layers are shown in Fig.~\ref{suppfig:network_architecture}-a. To address the loss-imbalance problem, we develop a multi-scale motion refinement strategy to synthesize finer motions and animate images from the resolution of $8\times8$ to $256\times256$ in a progressive generation manner, as shown in Fig.~\ref{suppfig:network_architecture}-b. The structure of Portrait Enhancer is depicted in Fig.~\ref{suppfig:network_architecture}-c. Detailed network blocks are presented in Fig.~\ref{suppfig:network_blocks}.

\subsection{Training Losses}
\label{subsec:training_losses}
We adopt multi-scale perceptual loss~\cite{siarohin2019first} and lip-sync loss~\cite{prajwal2020lip} to train our video generation model.

\noindent\textbf{Perceptual Loss.} Given the groundtruth frames ($I_1,...,I_T$) and synthesized frames ($\hat{I}_1, ..., \hat{I}_T$), and the feature extractor $\phi$. The perceptual loss can be written as follows: 
\begin{equation}
    \mathcal{L}_{perc}(\hat{I}_{1:T}, I_{1:T}) = \frac{1}{T}\sum_{t=1}^{T}\sum_{i=1}^{l}\lambda_{p}\|\phi_l(\hat{I}_t), \phi_l(I_t)\|_1,
\end{equation}
where $\lambda_p=10$. 

\noindent\textbf{Lip-Sync Loss.} To calculate the lip-sync loss, we first follow Wav2Lip~\cite{prajwal2020lip} and retrain a lip synchronization discriminator on our used dataset. The discriminator takes 5 continuous lip images and the corresponding mel-spectrogram as input, and employs a visual encoder and an audio encoder to extract two feature vectors (denoted by $f_v$ and $f_a$) from the cropped mouth images (denoted by $\hat{I}^{mouth}$) and mel-spectrogram, respectively. Therefore, we obtain the cosine similarity of these two embeddings as:
\begin{equation}
    C(f_v, f_a) = \frac{f_v\cdot f_a}{\max(\|f_v\|_2\|f_a\|_2, \epsilon)}, \epsilon=1e^{-7}.
\end{equation}
Therefore, the lip-sync loss can be written as:
\begin{equation}
    \mathcal{L}_{sync}=-\frac{1}{N}\sum_{i=1}^{N}\log(C(f^{i}_v, f^{i}_a)),
\end{equation}
where $i$ indicates $i$-th sample in a mini-batch and $N$ is the batch-size. 

In stage 1, we employ two losses to train the CoarseNet: multi-scale perceptual loss and lip-sync loss. The multi-scale perceptual loss is utilized to measure the perceptual distance between warped images and ground truth. We select the resolutions of $256\times256$ and $128\times128$ for loss measurement. 
Due the the loss imbalance problems, we emphasize the importance of the mouth region and penalize the reconstruction errors between the generated mouth images and the ground truth frames. The total loss for CoarseNet can be written as:
\begin{equation}
\begin{split}
    \mathcal{L}_{coarse} = \mathcal{L}_{sync} & + 0.5\mathcal{L}_{perc}(\hat{I}_{1:T}, I_{1:T})) \\
    & + 0.5\mathcal{L}_{perc}(\hat{I}^{mouth}_{1:T}, I^{mouth}_{1:T})
\end{split}
\end{equation}

In stage 2, we add lip-sync discrimination loss for more accurate lip motion refinement. The total loss for RefineNet can be written as:
\begin{equation}
\begin{split}
    \mathcal{L}_{refine} = \mathcal{L}_{sync} & + 0.5\mathcal{L}_{perc}(\hat{I}_{1:T}, I_{1:T}) \\
    & + 0.5\mathcal{L}_{perc}(\hat{I}^{mouth}_{1:T}, I^{mouth}_{1:T}).
\end{split}
\end{equation}
In stage 3, we optimize the Portrait Enhancer using only the perceptual loss. To optimize this network, we fixed the motion synthesis network trained in the first two stages, and then calculated a multi-scale perceptual reconstruction loss function based on the synthesis results and the ground truth image, resulting in an ultimate high-fidelity synthesized image. This approach ensures a seamless integration of the synthesized elements with the original content.

\begin{table}[h]
    \centering
    \small
    \setlength{\tabcolsep}{1mm}
    \begin{tabular}{l|l}
    \toprule
    Dataset  &  Target Name \\
    \hline
    \multirow{6}{*}{HDTF-Test} & AdamSchiff, AnnMarieBuerkle, AustinScott0 \\
        & BobCasey1, BobCorker, CoryGardner0 \\
        & DickDurbin, DuncanHunter, FredUpton \\
        & GeoffDavis, JoeHeck1, JohnHoeven \\
        & MichaelBennet, ReneeEllmers, RoyBlunt \\
        & ErikPaulsen, BrianSchatz2, BrianSchatz1 \\
        & CoryGardner1, JimInhofe, DougJones, JonKyl \\
        & CatherineCortezMasto, KayBaileyHutchison \\
        & AmyKlobuchar0, AmyKlobuchar1, AustinScott2 \\ 
        & CathyMcMorrisRodgers, TomPrice, JoePitts\\ 
    \midrule
    \multirow{4}{*}{MEAD-Train} & M005, M007, M009, M011, M012, M013, M019, \\
        &   M022, M024, M025, M028, M029, M030, M031, \\
        &   M032, M033, M037, M040, W009, W011, W014, \\
        &   W015, W016, W019, W028, W029, W035, W036  \\
    \hline
    MEAD-Test &  M003, W037 \\
    \bottomrule
    \end{tabular}
    \caption{IDs of subjects used for evaluation and training (MEAD). The remaining videos of HDTF are used for training. }
    \label{supptab:target_ids_split}
\end{table}

\begin{figure*}[htbp]
    \centering
    \includegraphics[width=1\linewidth]{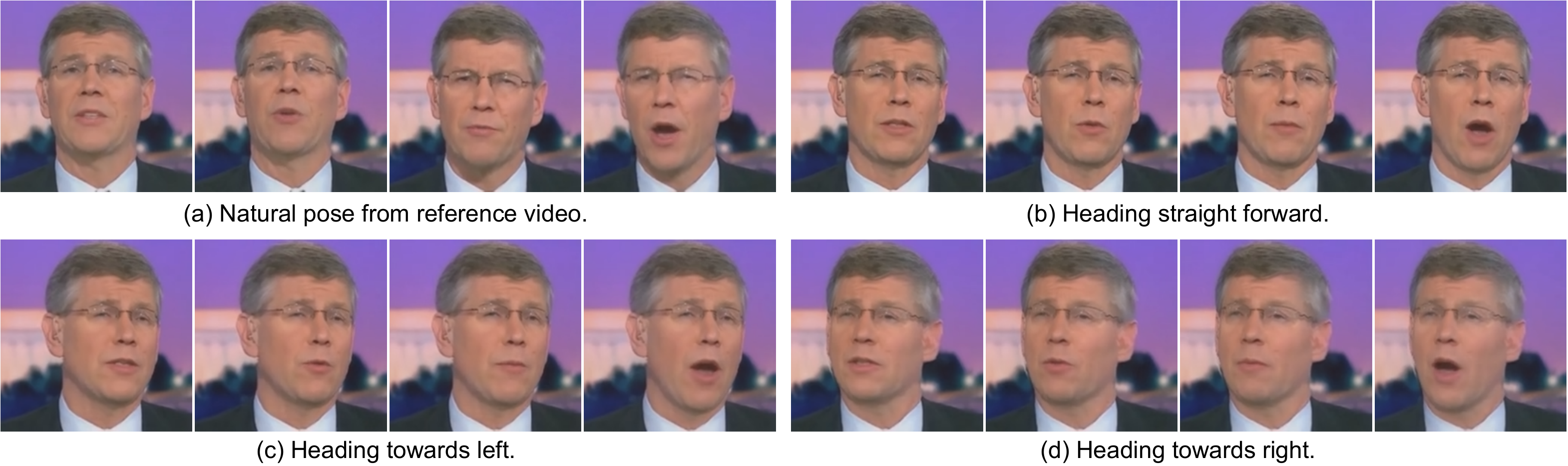}
    \caption{Talking head generation using different poses or audio. }
    \label{suppfig:audio_pose_editing}
\end{figure*}

\begin{figure*}[ht]
    \centering
    \includegraphics[width=1\linewidth]{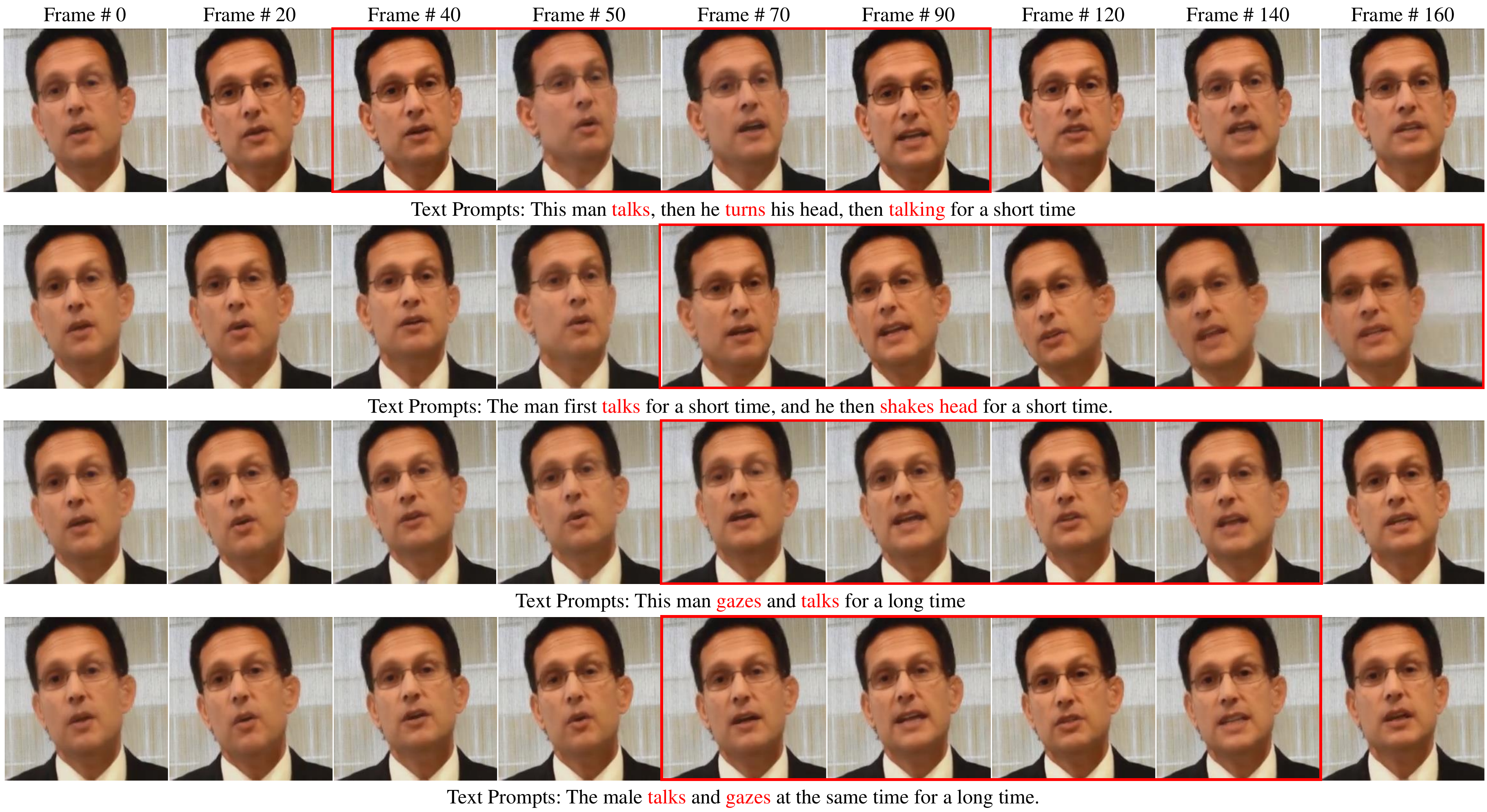}
    \caption{Talking head generation using the poses generated from the same audio but different text prompts. Key head movements are marked in red boxes. }
    \label{suppfig:text2pose_generation_text_variance}
\end{figure*}

\subsection{Testing Samples}
We sampled 30 objects from the HDTF~\cite{zhang2021flow} dataset and 2 subjects from MEAD~\cite{kaisiyuan2020mead} dataset for evaluation. The target names are shown in Table~\ref{supptab:target_ids_split}. 

\section{More Experiments}

\subsection{User Study}
Human evaluation is an essential subjective metric for quality evaluation in face video generation. We perform a user study to evaluate the generalization performance of our approach and previous methods. Specifically, we generate 28 videos using different methods, audio, and source images. During the study, we invited 5 experienced users to rate the videos' naturalness and lip-sync accuracy using 5-level scores ranging from 1 to 5 (Higher is better). The statistical results are shown in Table~\ref{table:user_study}. As can be seen, 
our method outperforms other competing methods in lip-sync quality and overall video naturalness, demonstrating the superiority of our approach.

\begin{table}[!htbp]
    \centering
    \setlength{\tabcolsep}{1.2mm}
    \begin{tabular}{l|cc}
    \toprule
     Method   & Naturalness$\uparrow$    &   Lip-Sync$\uparrow$ \\
     \hline
     Wav2Lip~\shortcite{prajwal2020lip}  &  2.25 & 4.00   \\
     PC-AVS~\cite{zhou2021pose}     &  3.35  & 4.05 \\
     SadTalker~\cite{zhang2023sadtalker}  & 3.45     & 3.30   \\
     IP\_LAP~\cite{zhong2023iplap}  & 2.20 & 2.65 \\
     PD-FGC~\cite{wang2022pdfgc} &   2.85     &   4.10    \\
     Hallo~\cite{xu2024hallo}  &    4.00    &   4.25   \\
     PoseTalk (Ours)       &  \textbf{4.05}   & \textbf{4.35} \\
    \bottomrule
    \end{tabular}
    \caption{User study on video generation quality. }
    \label{table:user_study}
\end{table}

\begin{figure*}[ht]
    \centering
    \includegraphics[width=1\linewidth]{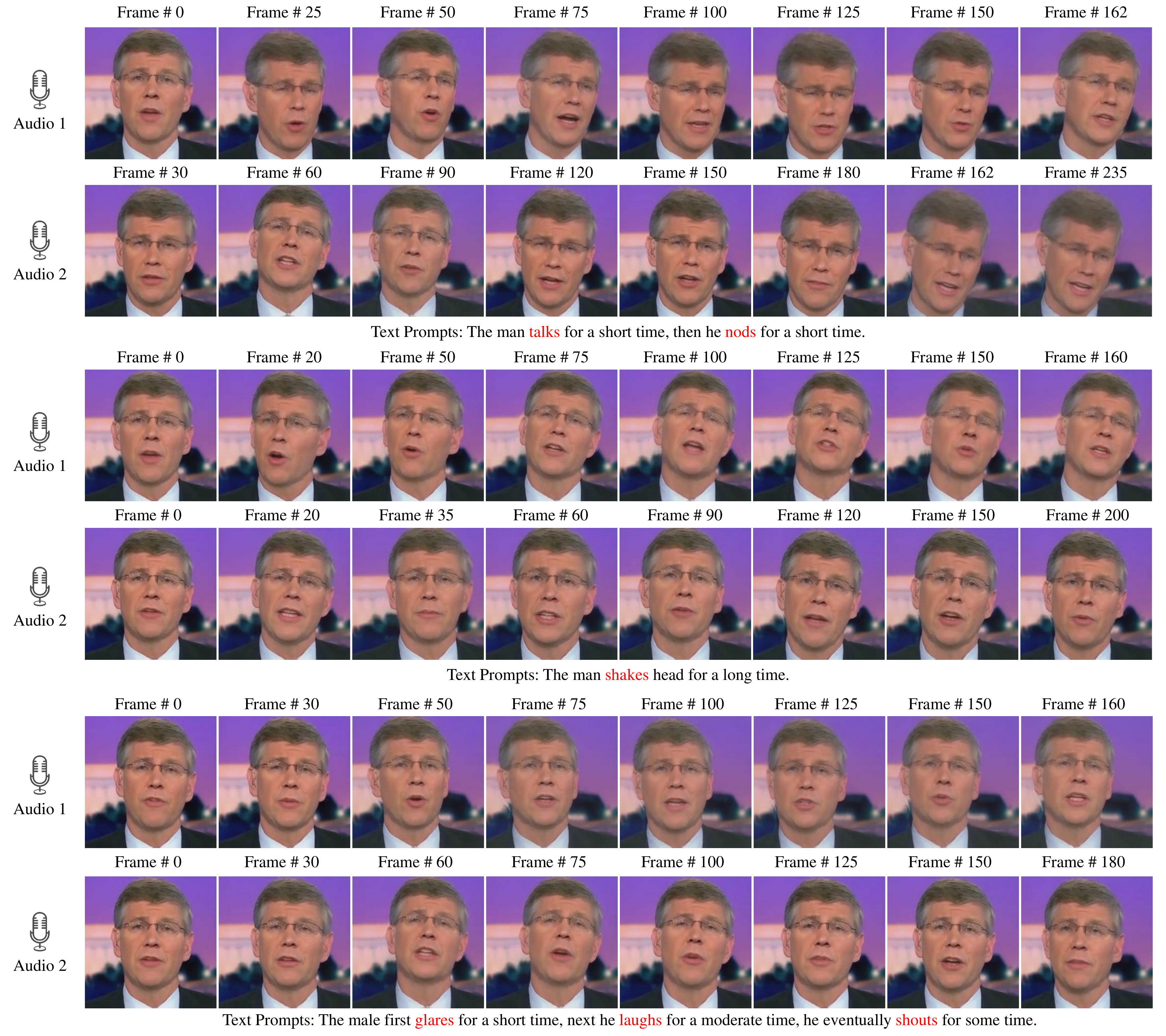}
    \caption{Talking head generation using the poses generated from the same text prompts but different audios. }
    \label{suppfig:text2pose_generation_audio_variance}
\end{figure*}

\subsection{Controllable Motion Synthesis}
Since we employ the disentangle poses and audio to control the motion details, the model is capable of modifying single motion conditions. Fig.~\ref{suppfig:audio_pose_editing} shows four groups of examples in which we generate videos under different poses. 
In Fig.~\ref{suppfig:text2pose_generation_text_variance}, we visualize the predicted poses from the text description and audio. As can be found, the generated videos vividly depict the action of `turns', `shakes', or 'gazes. 
We proceed to investigate the impact of audio on pose generation and show the results in Fig.~\ref{suppfig:text2pose_generation_audio_variance}. As can be found, our approach is capable of producing talking videos with changeable head poses influenced by input audio. 
We recommend readers watch demo videos in the supplementary file or browse our demo page \textcolor{blue}{\url{https://junleen.github.io/projects/posetalk}} for better observation.

\begin{figure}[ht]
    \centering
    \includegraphics[width=1\linewidth]{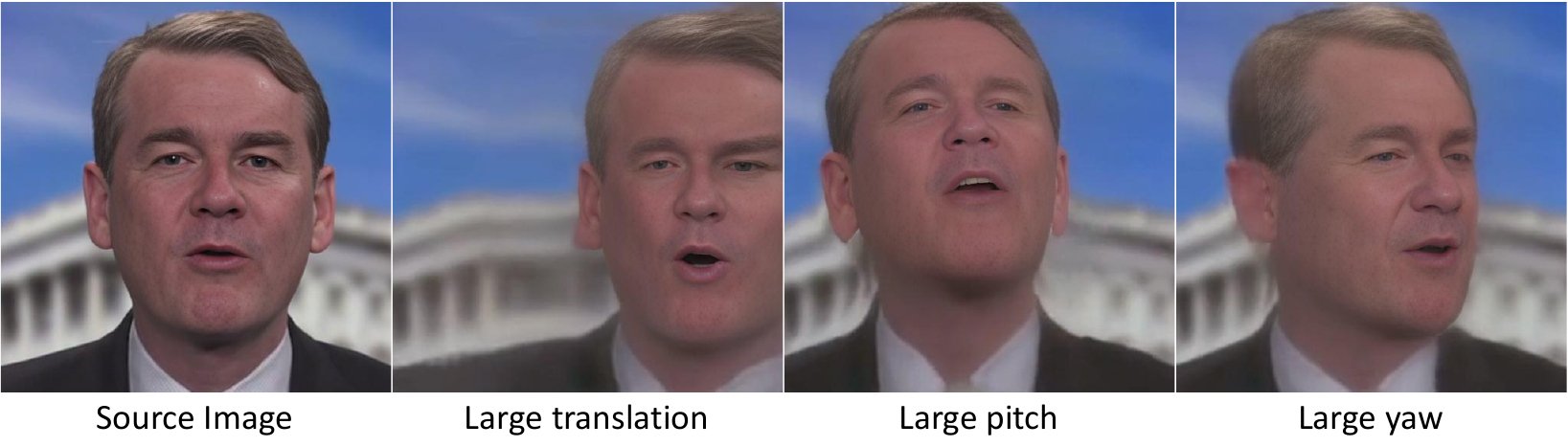}
    \caption{Failure cases. Limited by the pose diversities in the training dataset, our generator might produce blurry artifacts under large poses. }
    \label{suppfig:limitations}
\end{figure}
\subsection{Discussions and Limitations}
\label{subsec:exp_discussion}

Images, audio, and languages are signals of different natures and this can be addressed by training models on large amounts of paired multimedia datasets~\cite{he2022masked}. Our work is inspired by the success of generative models in text-to-image/audio/video areas~\cite{rombach2022high,liu2023audioldm,videoworldsimulators2024}. Instead of attempting to construct a paired and large-scale multi-modal dataset (text, image, video), our approach fully utilizes off-the-shelf face video datasets. 
Thanks to our setting, we divide our goals into two tasks and conquer those goals one by one. Meanwhile, our approach can be used for mouth retalking by taking the poses and gazes from the original video, thereby allowing our model to modify only the mouth regions. 
Furthermore, our approach can be developed for video dubbing and language translation in movies. 
Compared to 0.5fps of Hallo~\cite{xu2024hallo}, our video generator can run over 100 fps on a single NVIDIA A100 GPU, thus enabling interactive talking head generation for live talks and video conferencing. 

Although our approach achieves competitive performance in balancing lip-synchronization quality and pose generation, it does not come without limitations. For example, the video generator might suffer from extreme poses (see Fig.~\ref{suppfig:limitations}), which is mainly introduced by the limited pose diversity of the talking videos in HDTF and MEAD. A proper solution for this issue should be involving more in-the-wild talking videos with diverse pose styles for training. In addition, the text condition in CelebV-Text~\cite{yu2023celebv-text} cannot fully describe the detailed motions, such as ``\textit{The man rotates his head to the right by 45$^\circ$} and ". Future explorations on more fine-grained text descriptions should be required. 

\end{document}